\documentclass[sn-apa]{sn-jnl}

\usepackage{multicol}
\usepackage{natbib}
\usepackage{tabularx}
\newcolumntype{b}{X}
\newcolumntype{s}{>{\hsize=.4\hsize}X}
\usepackage{multirow}
\usepackage{url}
\usepackage{amsmath}
\usepackage{array}
\usepackage{tabu}
\usepackage{bm}
\usepackage[caption=false]{subfig}
\usepackage{graphicx}
\usepackage{makecell}
\usepackage{diagbox}
\usepackage{xcolor}
\usepackage{algorithm}
\usepackage{algpseudocode}
\algnewcommand{\LineComment}[1]{\State \(\triangleright\) #1}
\algnewcommand\Var[1]{\mbox{\ttfamily\detokenize{#1}}}
\usepackage{etoolbox}

\usepackage{xcolor}

\jyear{2021}%

\begin{document}

\title[Rumor Classification by Multimodal Fusion and Ensemble Learning]{Rumor Classification through a Multimodal Fusion Framework and Ensemble Learning}


\author*[1]{\fnm{Abderrazek} \sur{Azri}}\email{a.azri@univ-lyon2.fr}
\author[1]{\fnm{C\'ecile} \sur{Favre}}\email{cecile.favre@univ-lyon2.fr}
\author[1]{\fnm{Nouria} \sur{Harbi}}\email{nouria.harbi@univ-lyon2.fr}
\author[1]{\fnm{J\'er\^ome} \sur{Darmont}}\email{jerome.darmont@univ-lyon2.fr}
\author[2]{\fnm{ Camille} \sur{ No\^us}}\email{camille.nous@cogitamus.fr}

\affil[1]{\orgdiv{Université de Lyon, Lyon 2}, \orgname{UR ERIC},\\ \orgaddress{\street{5 avenue Pierre Mend\`es France}, \postcode{69676} \city{Bron Cedex}, \country{France}}}

\affil[2]{\orgdiv{Université de Lyon, Lyon 2}, \orgname{Laboratoire Cogitamus},  \country{France}}



\abstract{The proliferation of rumors on social media has become a major concern due to its ability to create a devastating impact. Manually assessing  the veracity of social media messages is a very time-consuming task that can be much helped by machine learning. Most message veracity verification methods only exploit textual contents and metadata. Very few take both textual and visual contents, and more particularly images, into account. Moreover, prior works have used many classical machine learning models to detect rumors. However, although recent studies have proven the effectiveness of ensemble machine learning approaches, such models have seldom been applied.  Thus, in this paper, we propose a set of advanced image features that are inspired from the field of image quality assessment, and introduce the Multimodal fusiON framework to assess message veracIty in social neTwORks (MONITOR), which exploits all message features by exploring various machine learning models. Moreover, we demonstrate the effectiveness of ensemble learning algorithms for rumor detection by using five metalearning models. Eventually, we conduct extensive experiments on two real-world datasets. Results show that MONITOR outperforms state-of-the-art machine learning baselines and that  ensemble models significantly increase MONITOR's performance.}




\maketitle

\section{Introduction}\label{sec:introduction}

After more than two decades of existence, social media platforms have attracted a large number of users. They enable the diffusion of information in real-time, albeit regardless of its credibility, for two main reasons. First, there is a lack of a means to verify the veracity of contents transiting on social media. Second, users often publish messages without verifying information validity and reliability. Consequently, social networks, and particularly microblogging platforms, are a fertile ground for spreading rumors. 

Widespread rumors can pose a threat to the credibility of social media and cause harmful consequences in real life. Thus, the automatic assessment of information credibility on microblogs that we focus on is crucial to provide decision support to, e.g., fact checkers. This task requires to verify the truthfulness of messages related to a particular event and return a binary decision  
stating whether the message is authentic.

In the literature, most automatic rumor detection approaches address the task as a classification problem. They generally extract features from two aspects of messages: textual content \citep{perez-rosas-etal-2018-automatic} and social context \citep{wu2018tracing}. However, the multimedia content of messages, particularly images that present a significant set of features, are little exploited.

In this paper, we second the hypothesis that the use of image properties is important in rumor verification. Images indeed play a crucial role in the news diffusion process. For example, in the dataset collected by \cite{jin2017novel}, the average number of messages with an attached image is more than eleven times that of plain text messages. 

\begin{figure*}[htp]
	\centering
	\subfloat[Black clouds in New York City before Sandy!!!\label{a}]{%
		\includegraphics[width=0.35\textwidth]{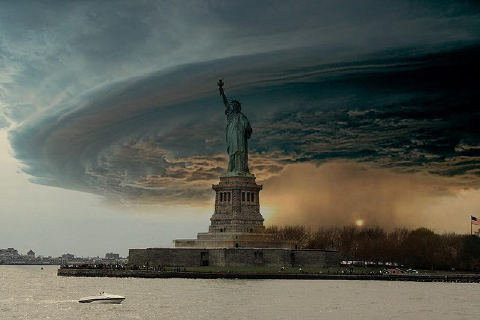}%
	}\hfil
	\subfloat[NepalEarthquake 4Years old boy protect his little sister. make me feel so sad\label{b}]{%
		\includegraphics[width=0.25\textwidth]{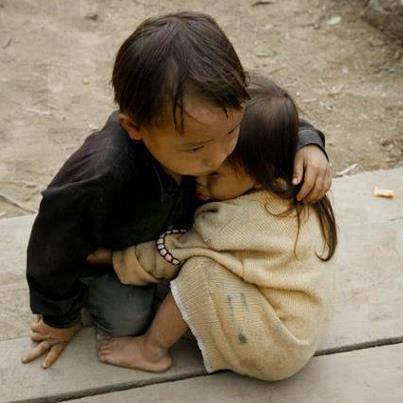}%
	}
	\caption{Two sample rumors posted on Twitter}
	\label{Fig1}
\end{figure*}

Figure~\ref{Fig1} shows two sample rumors posted on Twitter. In Figure~\ref{a}, it is hard to assess veracity from the text, but the likely-manipulated image hints at a rumor. In Figure~\ref{b}, it is hard to assess veracity from both the text or the image because the image has been taken out of its original context. 

Furthermore, most of the literature focuses on features to train a wide range of machine learning \citep{volkova2018misleading} and deep learning \citep{wang2018eann} methods. However, although recent studies demonstrate the effectiveness of ensemble learning \citep{gutierrez2020fake}, such models are not
applied for rumor detection.

Based on the above observations, we aim to leverage all the modalities of microblog messages for verifying rumors, that is, features extracted from the textual and social context content of messages, and up to now unused visual and statistical features derived from images. Consequently, all types of features must be fused to allow a supervised machine learning classifier to evaluate the credibility of messages. Moreover, motivated by the recent research on ensemble learning to classification problems \citep{pang2016early}, we design various metalearning models to investigate the performance of ensemble learning for rumor classification.

Our contribution is threefold. First, we propose the use of a set of image features inspired from the field of Image Quality Assessment (IQA) and we show that they contribute very effectively to the verification of message veracity. These metrics estimate the rate of noise and  quantify the amount of visual degradation of any type in an image. They are proven to be good indicators for detecting fake images, even those generated by advanced techniques such as Generative Adversarial Networks (GANs) \citep{goodfellow2014generative}.  To the best of our knowledge, we are the first to systematically exploit this type of image features to check the veracity of microblog posts. 

Second, we detail the Multimodal fusiON framework to assess message veracIty in social neTwORks (MONITOR) \citep{azri2021monitor}, which exploits all types of message features and leverages four machine learning models that provide explainability and interpretability about the taken decisions. 

Third, we demonstrate the benefit of ensemble learning, by developing five metalearning models (soft and weighted average voting, stacking, blending, and super learner ensemble) that exploit the above four machine learning models, and we compare their performance with MONITOR's. To the best of our knowledge, we are the first to apply metalearning models for tackling the rumor detection task.  

Eventually, we conduct extensive experiments two real-world datasets to show the effectiveness of our rumor detection approach. MONITOR indeed outperforms all state-of-the-art machine learning baselines with an accuracy and F1-score of up to 96\% and 89\% on the MediaEval benchmark \citep{boididou2015verifying} and the FakeNewsNet dataset \citep{shu2018fakenewsnet}, respectively. Furthermore,  all metalearning algorithms notably increase MONITOR's performance.    

The remainder of this paper is organized as follows. In Section~\ref{sec:relatedworks}, we review all the research related to our problem. In Section~\ref{sec:framework}, we detail MONITOR and especially its feature extraction and selection. In Section~\ref{sec:experiments}, we present and comment on the experimental results that we achieve with respect to state-of-the-art methods. In Section~\ref{sec:ensemble}, we investigate and discuss  the performance of ensemble models. Finally, in Section~\ref{sec:conclusion}, we conclude this paper and outline future research. 

\section{Related Works}
\label{sec:relatedworks}

Related work can be divided into the following categories: 
\begin{enumerate}
    \item non-image features and image features that are essential for checking the veracity of microblog posts,
     \item background information regarding ensemble learning models and their usage for rumor classification.  
\end{enumerate}

\subsection{Non-image Features}
Studies in the literature present a wide range of non-image features. These features may be divided into two subcategories, textual features and social context features. To classify a message as fake or real, \cite{castillo2011information} capture prominent statistics in tweets, such as count of words, capitalized characters and punctuation. Beyond these features, lexical words expressing specific semantics or sentiments  are also counted. Many sentimental lexical features are proposed  \citep{kwon2013prominent}, which utilize a sentiment tool called the Linguistic Inquiry and Word Count (LIWC) to count words in meaningful categories. 

Other works exploit syntactic features, such as the number of keywords, the sentiment score or polarity of the sentence. Features based on topic models are used to understand messages and their underlying relations within a corpus. \cite{wu2015false} train a Latent Dirichlet Allocation model \citep{blei2003latent} with a defined set of topic features to summarize semantics for detecting rumors.  

The social context describes the propagating process of a rumor \citep{shu2018understanding}. Social network features are extracted by constructing specific networks, such as diffusion \citep{kwon2013prominent} or co-occurrence networks \citep{ruchansky2017csi}.

Recent approaches detect fake news based on temporal-structure features. \cite{kwon2017rumor} studied the stability of features over time and found that, for rumor detection, linguistic and user features are suitable for early-stage, while structural and temporal features tend to have good performance in the long-term stage.
\subsection{Image Features}
Although images are widely shared on social networks, their potential for verifying the veracity of messages in microblogs is not sufficiently explored.  \cite{morris2012tweeting} assume that the user's profile image has an important impact on information credibility. 
Images attached in messages bear very basic features.  \cite{wu2015false} define a feature called ``has multimedia''  to mark whether the tweet has any picture, video or audio attached. \cite{gupta2013faking} propose a classification model to identify fake images on Twitter during Hurricane Sandy. However, their work is still based on textual content features. 

To automatically predict whether a tweet that shares multimedia content is fake or real,  \cite{boididou2015verifying} propose the Verifying Multimedia Use (VMU) task. Textual and image forensics \citep{li2014segmentation} features are used as baseline features for this task. They conclude that Twitter media content is not amenable to image forensics and that forensics features do not lead to consistent VMU improvement \citep{boididou2018detection}. 

\subsection{Ensemble Learning}

Ensemble learning refers to the generation and combination of multiple inducers to solve a particular machine learning task. The intuitive explanation for the ensemble methodology stems from human nature. Often, decision making by a group of individuals results in more accurate, useful or correct outcome than a decision made by any one member of the group. This is generally referred to as the wisdom of the crowd \citep{surowiecki2005wisdom}. Using ensemble learning, the performance of poorly performing classifiers can be improved by creating, training and combining the output of multiple classifiers and thus result in a more robust classification. There are three main approaches for developing an ensemble learner \citep{zhang2012ensemble}:
\begin{itemize}
	\item \textit{boosting} uses homogeneous-base models trained sequentially;
	\item \textit{bagging} (Bootstrap AGGregatING) uses homogeneous-base models trained in parallel; 
	\item \textit{stacking} uses mostly heterogeneous-base models trained in parallel and combined using a metamodel.
\end{itemize}
By averaging (or voting) the output produced by the pool of classifiers, ensemble methods provide better predictions and avoid overfitting. Another reason that contributes to the better performance of ensemble learning is its ability in escaping from local minimums. By using multiple models, the search space becomes wider and the chance for finding a better output becomes higher \citep{sagi2018ensemble}.

Recently ensemble learning methods have shown good performance in various applications, including solar irradiance prediction \citep{lee2020reliable}, slope stability analysis \citep{pham2021ensemble}, natural language processing \citep{sangamnerkar2020ensemble}, malware detection \citep{gupta2020improving}, 
COVID-19 detection \citep{singh2021novel}, movie success detection \citep{lee2018predicting} and blood donors detection \citep{kauten2021predicting}. Compared to other applications, rumor classification using ensemble learning techniques has been very little studied. 

\cite{kaur2020automating} propose a multilevel voting model for the fake news detection task. The study concludes that the proposed model outperforms both individual machine learning and ensemble
learning models. To address the multiclass fake news detection problem,  \cite{kaliyar2019multiclass} use gradient boosting ensemble techniques and compare their
performance with several individual machine learning models. Results demonstrate the effectiveness of the ensemble framework compared to existing benchmark performance.  Finally, \cite{al2019ensemble} find that the bagging approach provides superior performance than Support Vector Machines (SVMs), Multinomial Naïve Bayes (MNB) and Random Forest to detect fake news.
\section{MONITOR}

\label{sec:framework}

Microblog messages contain rich multimodal resources, such as text contents, surrounding social context and attached images. Our focus is to leverage this multimodal information to determine whether a message is true or false. Based on this  idea, we propose a framework for verifying the veracity of messages. MONITOR's detailed description is presented in this section. 

\subsection{Multimodal Fusion Overview}

\begin{figure}[htbp]
	\centerline{\includegraphics[width=0.85\textwidth]{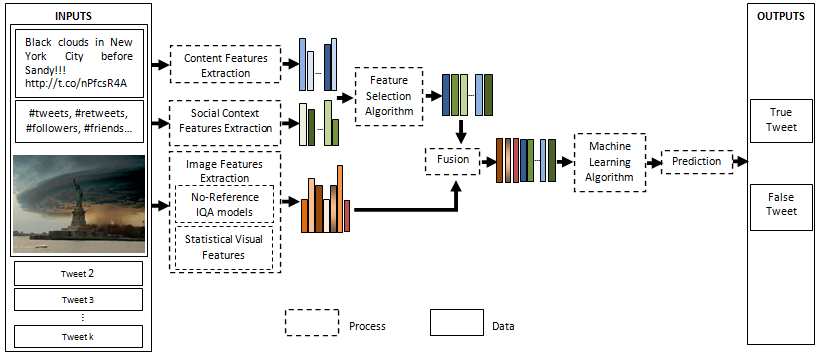}}
	\caption{Overview of MONITOR}
	\label{Fig2} 
\end{figure}

Figure~\ref{Fig2} shows a general overview of MONITOR, which works in two main stages. First, we 
extract several features from the message's text and the social context. Then, we apply a feature selection algorithm to identify relevant features, which form a first set of textual features. From the attached image, we derive statistics and efficient visual features inspired from the IQA field, which form a second set of image features. 
Second, we train a model by concatenating and normalizing the textual and image features sets to form a fusion vector. Several machine learning classifiers may learn from the fusion vector to distinguish the veracity of the message, i.e., real or fake.

\subsection{Feature Extraction and Selection}

To better extract features, we reviewed the best practices followed by information professionals, e.g., journalists, in verifying content generated by social network users. We based our thinking on relevant data from journalistic studies \citep{martin2014information} and the Verification Handbook \citep{silverman2014verification}. We define a set of features that are important to extract discriminating characteristics of rumors. These features are mainly derived from three principal aspects of news information: content, social context and visual content. The feature selection process is only applied to content and social context features sets to remove the irrelevant features that can negatively impact performance. Because our focus is the visual features set, we retain all these features in the learning process. 

\subsubsection{Message Content Features}  

Content features are extracted from the message's text. We extract characteristics such as the length of a tweet and the number of words. We also include statistics such as the number of exclamation and question marks, as well as binary features indicating the existence or not of emoticons. Furthermore, other features are extracted from the linguistics of a text, including the number of positive and negative sentiment words. Additional binary features indicate whether the text contains personal pronouns.

We also calculate  a readability score for each message using the Flesch Reading Ease method \citep{kincaid1975derivation}. The higher this score is, the easier the text is to read. Other features are extracted from the informative content provided by the specific communication style of the Twitter platform, such as the number of retweets, mentions~(@), hashtags~(\#) and URLs. 

\subsubsection{Social Context Features}  
The social context reflects the relationships between  different users. Therefore, social context features are extracted from the behavior of users and the propagation network. We capture several features from the users' profiles, such as the number of followers and friends, the number of tweets the user has authored, the number of tweets the user has liked and whether the user is verified by the social media. We also extract features from the  propagation tree that can be built from tweets and retweets, such as the depth of the retweet tree. Tables~\ref{Tab1} and \ref{Tab2}  describe the sets of content features and social context features extracted from each message.

\begin{table}[!ht]
	\centering
	\begin{minipage}[t]{0.40\linewidth}\centering
		\caption{Content features}
		\label{Tab1}
		\begin{tabular}{l}
			\hline
			\thead{Description}\\
			\hline
			\# of chars, words\\
			\# of (?), (!) mark \\
			\# of uppercase chars \\
			\# of positive, negative words \\
			\# of mentions, hashtags, URLs\\
			\# of happy, sad mood emoticon \\
			\# of 1\textsuperscript{st}, 2\textsuperscript{nd}, 3\textsuperscript{rd} order pronoun\\
			Readability score\\
			\hline
		\end{tabular}
	\end{minipage}\quad%
	\begin{minipage}[t]{0.40\linewidth}\centering
		\caption{Social context features}
		\label{Tab2}
		\begin{tabular}{l}
			\hline
			\thead{Description}\\
			\hline
			\# of followers, friends, posts\\
			Friends/followers ratio, times listed \\
			\# of retweets, likes\\
			The user shares a homepage URL\\
			The user has a profile image\\
			The user has a verified account\\
			\# of tweets the user has liked\\
			\hline
		\end{tabular}
	\end{minipage}
\end{table}

To improve the performance of MONITOR, we apply a feature selection algorithm on the feature sets listed in Tables~\ref{Tab1} and \ref{Tab2}. The details of the feature selection process are discussed in Section~\ref{sec:experiments}.

\subsubsection{Image Features}  
\label{sec:imgfeat}

To differentiate between false and real images in messages, we propose to exploit visual content features and visual statistical features that are extracted from the joined images.

\paragraph{Visual Content Features} Usually, a news consumer decides the image veracity based on his subjective perception, but how do we quantitatively represent the human perception of the quality of an image? The quality of an image means the amount of visual degradations of all types present in an image, such as noise, blocking artifacts, blurring, fading and so on. 

The IQA field aims to quantify human perception of image quality by providing an objective score of image degradations based on computational models \citep{maitre2017photon}. Such degradations are introduced during different processing stages, such as image acquisition, compression, storage, transmission and decompression. Inspired by the potential relevance of IQA metrics in our context, we use these metrics in an original way, for a purpose different from what they were created for. More precisely, we hypothesize that the quantitative evaluation of the quality of an image can be useful for veracity detection. 

IQA is mainly divided into two areas of research:  full-reference evaluation and no-reference evaluation. Full-reference algorithms compare the input image against a pristine reference image with no distortion. In no-reference algorithms, the only input is the image whose quality is to be measured. 
In our case, we do not have the original version of the posted image. Therefore, the approach that is fitting to our context is no-reference evaluation. We use three no-reference algorithms that have been demonstrated to be highly efficient: the Blind/Referenceless Image Spatial Quality Evaluator (BRISQUE)  by \cite{mittal2011blind}, the Naturalness Image Quality Evaluator (NIQE) by \cite{mittal2012making} and the Perception based Image Quality Evaluator (PIQE) by  \cite{venkatanath2015blind}.

\begin{figure}[htp]
	\centering
	\captionsetup[subfloat]{position=bottom,labelformat=empty, justification=centering}
	
	\subfloat[Original image 13.7215\label{}]{%
		\includegraphics[width=0.20\textwidth]{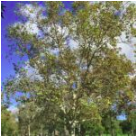}%
	}\quad
	\subfloat[JPEG compressed 22.6603\label{}]{%
		\includegraphics[width=0.20\textwidth]{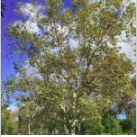}%
	}\quad
	\subfloat[Gaussian noise 28.5840\label{}]{%
		\includegraphics[width=0.20\textwidth]{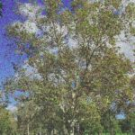}%
	}\quad
	\subfloat[Median blur 41.5620\label{}]{%
		\includegraphics[width=0.20\textwidth]{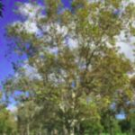}%
	}
	\caption{BRISQUE score  computed for a natural image and its distorted versions}
	\label{Fig3}
	
\end{figure}

For example, Figure~\ref{Fig3} displays the  BRISQUE score computed for a natural image and its distorted versions (compression, noise and blurring distortions). The BRISQUE score is a  non-negative scalar in the range [1, 100]. Lower values of the score reflect a better perceptual image quality.

\begin{figure}[htp]
	\centering
	\captionsetup[subfloat]{position=bottom,labelformat=empty,justification=centering}
	\subfloat[Real image 17.7778\label{}]{%
		\includegraphics[width=0.18\textwidth]{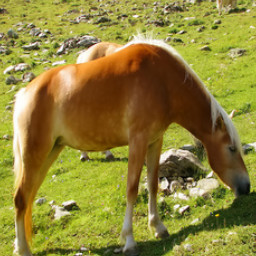}%
	}\quad
	\subfloat[Fake image 22.0260\label{}]{%
		\includegraphics[width=0.18\textwidth]{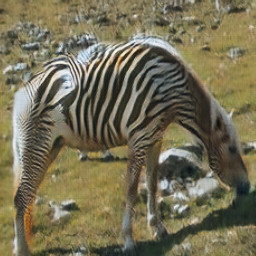}%
	}\quad
	\subfloat[Real image 12.5000\label{}]{%
		\includegraphics[width=0.18\textwidth]{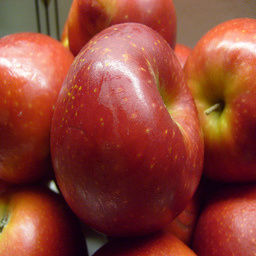}%
	}\quad
	\subfloat[Fake image 22.5279\label{}]{%
		\includegraphics[width=0.18\textwidth]{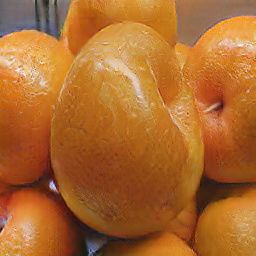}%
	}
	\caption{BRISQUE score computed for real and fake GANs images}
	\label{Fig4}
	
\end{figure}

No-reference IQA metrics are also good indicators for other types of image modifications, such as GAN-generated images. These techniques allow modifying the context and semantics of images in a very realistic way. Unlike many image analysis tasks, where both reference and reconstructed images are available, images generated by GANs may not have any reference image. This is the main reason for using no-reference IQA for evaluating this type of fake images. Figure~\ref{Fig4} displays the BRISQUE score computed for real and fake images generated by image-to-image translation based on GANs \citep{CycleGAN2017}.

\paragraph{Statistical Features} From attached images, we define four statistical features from two aspects.
\begin{itemize}
\item \textit{Number of images:} A user can post one, several or no images. To denote this feature, we count the total number of images in a rumor event and the ratio of posts containing more then one image.
\item \textit{Spreading of images:} During an event, some images are very replied and generate more comments than others. The ratio of such images is calculated to indicate this feature. Table~\ref{tab3} illustrates the description of our visual and statistical features. We use all of these features in the learning process.   
\end{itemize}

\begin{table}[htbp]
\begin{center}
	\caption{Description of image features}
	
		\begin{tabular}{l|l|l}
			\hline
			\thead{Type}&\thead{Feature}&\thead{Description}\\
			\hline
			\multirow{3}{*}{Visual}&BRISQUE&BRISQUE score of a given image\\
			&PIQE&PIQE score of a given image\\
			features&NIQE&NIQE score of a given image\\
			\hline
			\multirow{3}{*}{Statistical}&Count\_Img&Number of all images in a news event\\
			&Ratio\_Img1&Ratio of the multi-image tweets in all tweets\\
			features&Ratio\_Img2&Ratio of image number to tweet number\\
			&Ratio\_Img3&Ratio of the most widespread image in all distinct images\\
			\hline
		\end{tabular}
		\label{tab3}
	\end{center}
\end{table}

\subsection{Model Training}
So far, we have obtained a first set of relevant textual features through a feature selection process. We have also a second set of image features composed of statistical and visual features. These two sets of features are scaled, normalized and concatenated to form the multimodal representation of a given message, which is learned by a supervised classifier. Several learning algorithms can be implemented fore message veracity  classification. We investigate the algorithms that provide the best performance in Section~\ref{sec:experiments}.

\section{Regular Machine Learning Experiments}
\label{sec:experiments}
In this section, we conduct extensive experiments on two  public datasets. First, we present statistics about the datasets we use. Then, we describe the experimental settings: a brief review of state-of-the-art features for news verification and a selection of the best of these textual features as baselines. Finally, we present experimental results and analyze the features to achieve insights with MONITOR.

\subsection{Datasets}
To evaluate MONITOR's performance, we conduct experiments on two well-established public datasets for rumor detection. The detailed statistics of these two datasets are listed in Table~\ref{tab4}.

\subsubsection{MediaEval}
MediaEval \citep{boididou2015verifying} is collected from Twitter and includes all three characteristics: text, social context and images. It is designed for message-level verification. The dataset has two parts: a development set containing about 9,000 rumor and 6,000 non-rumor tweets from 17 rumor-related events; a test set containing about 2,000 tweets from another batch of 35 rumor-related events. We remove tweets without any text nor image, thus obtaining a final dataset including 411 distinct images associated with 6,225 real and 7,558 fake tweets, respectively.

\subsubsection{FakeNewsNet}
FakeNewsNet  \citep{shu2018fakenewsnet} is one of the most comprehensive fake news detection benchmark. Fake and real news articles are collected from the fact-checking websites PolitiFact and GossipCop. Since we are particularly interested in images in this work, we extract and exploit the image information of all tweets. To keep the dataset balanced, we randomly choose 2,566 real and 2,587 fake news events. After removing tweets without images, we obtain 56,369 tweets and 59,838 images.

\begin{table}[!ht]
	\caption{MediaEval and FakeNewsNet statistics}\label{tab4}
	\centering
	\begin{tabular}{c|c|cc|c}
		\hline
		\multirow{2}{*}{Dataset}&\multirow{2}{*}{Set}&\multicolumn{2}{c|}{Tweets}&\multirow{2}{*}{Images}\\
		&&Real&Fake&\\
		\hline 
		MediaEval&Training set&5,008&6,841&361\\
		&Testing set&1,217&717&50\\
		\hline
		FakeNewsNet&Training set&25,673&19,422&47,870\\
		&Testing set&6,466&4,808&11,968\\	
		\hline
	\end{tabular}
\end{table}

\subsection{Experimental Settings}


\subsubsection{Baseline Features} We compare the effectiveness of our feature set with the best textual features from the literature. First, we adopt the 15 best features extracted by \cite{castillo2011information} to analyze the information credibility of news propagated through Twitter. We also collect a total of 40 additional textual features from the literature \citep{gupta2013faking,gupta2012evaluating,kwon2013prominent,wu2015false}, which are extracted from text content, user information and propagation properties (Table~\ref{Tab5}).

\begin{table}[!ht]
	\centering
	\begin{minipage}[t]{0.46\linewidth}\centering
		\caption{Features from the literature}
		\label{Tab5}
		\begin{tabular}{l}
			\hline
			\thead{Feature}\\
			\hline
			Fraction of (?), (!) Mark, \# of messages\\
			Average \# of words, char lengths\\
			Fraction of 1\textsuperscript{st},  2\textsuperscript{nd}, 3\textsuperscript{rd} pronouns\\
			Fraction of URLs, @, \#\\
			Count of distinct URLs, @, \#\\
			Fraction of popular URLs, @, \#\\
			The tweet includes pictures\\
			Average sentiment score\\
			Fraction of positive and negative tweets\\
			\# of distinct people, loc, org\\
			Fraction of people, loc, org\\
			Fraction of popular people, loc, org\\
			\# of Users, fraction of popular users\\
			\# of followers, followees, posted tweets\\
			The user has a Facebook link\\
			Fraction of verified users, org\\
			\# of comments on the original message\\
			Time between original message and repost\\
			\hline
		\end{tabular}
	\end{minipage}\qquad%
	\begin{minipage}[t]{0.45\linewidth}\centering
		\caption{Best textual features selected 
		}
		\label{Tab6}
		\begin{tabular}{l|l}
			\hline
			\thead{MediaEval}&\thead{FakeNewsNet}\\
			\hline
			Tweet\_Length&Tweet\_Length \\
			Num\_Negwords&Num\_Words\\
			Num\_Mentions&Num\_Questmark\\
			Num\_URLs&Num\_Upperchars\\
			Num\_Words&Num\_Exclmark\\
			Num\_Upperchars&Num\_Hashtags\\
			Num\_Hashtags&Num\_Negwords\\
			Num\_Exclmark&Num\_Poswords\\
			Num\_Thirdpron&Num\_Followers\\
			Times\_Listed&Num\_Friends\\
			Num\_Tweets&Num\_Favorites\\
			Num\_Friends&Times\_Listed\\
			Num\_Retweets&Num\_Likes\\
			Has\_Url&Num\_Retweets\\
			Num\_Followers&Num\_Tweets\\
			\hline
		\end{tabular}
	\end{minipage}
\end{table}

\subsubsection{Feature Sets} 

The features labeled \textit{Textual} are the best features selected among message content and social context features (Tables~\ref{Tab1} and \ref{Tab2}). We select them  with the information gain ratio method~\citep{karegowda2010comparative}, which helps select a subset of 15 relevant textual features with an information gain larger than zero (Table~\ref{Tab6}). 

The features labeled \textit{Image} are all the image features listed in Table~\ref{tab3}. The features labeled \textit{MONITOR} are the feature set that we propose, consisting of the fusion of textual and image feature sets. The features labeled \textit{Castillo} are the above-mentioned best 15 textual features. Eventually, the features labeled \textit{Wu} are the 40 textual features identified in literature. 


\subsubsection{Model Construction} 
We cannot know beforehand what model will be good for our problem or what configuration to use. By analyzing both datasets, we found that classes are partially linearly separable in some dimensions. Thus, we evaluate a mix of simple linear and non-linear algorithms. The best result are achieved by four supervised classification algorithms: Classification and Regression Trees (CART), $k$-Nearest Neighbors (KNN), Support Vector Machines (SVMs) and Random Forest (RF). Then, we  optimize the hyper-parameters of each model (Table \ref{tab7}) by testing multiple settings using the \textit{GridSearchCV} function from the  Python Scikit-Learn library~\citep{scikit-learn}. Subsequently, we perform training and validation for each model through a 5-fold cross-validation to obtain stable out-sample results. To implement the models, we again use scikit-learn. Note that, for MediaEval, we retain the same data split scheme. For FakeNewsNet, we randomly divide data into training and testing subsets with the ratio 0.8:0.2. Table~\ref{tab8} present the results of our experiments.

\begin{table}
	\caption{Hyper-parameters configuration space}\label{tab7}
	\centering
	\begin{tabular}{c|c|c|c}
		\hline
		\thead{Model}&\thead{Main hyper-parameters}&\thead{Type}&\thead{Search space}\\
		\hline
		CART &max\_depth&Discrete&[1,21]\\
		&criterion&Categorical&['gini','entropy']\\
		\hline
		KNN &n\_neighbors&Discrete&[1,21]\\
		\hline
		SVM&C&Discrete&[0.1,2.0]\\
		&$\gamma$ (RBF kernel)&Discrete&[0.1,1.0]\\
		&Kernel&Categorical&[‘linear’, ‘poly’, ‘rbf’,‘sigmoid’]\\
		\hline
		RF&n\_estimators&Discrete&[10,500]\\
		&max\_depth&Discrete&[3,20]\\
		\hline
	\end{tabular}
\end{table}

\begin{table}
	\caption{Performance of individual machine learning models}\label{tab8}
	\centering
	
	\begin{tabularx}{\textwidth}{bbssssssss}
		\hline
		&&\multicolumn{4}{c}{\textbf{MediaEval}}&\multicolumn{4}{c}{\textbf{FakeNewsNet}}\\
		\multirow{2}{*}{\textbf{Model}} & \multirow{2}{*}{\textbf{Features}} & &&&&&&&\\
		&  &\textbf{Acc} & \textbf{Prec}&\textbf{Rec}&  $\bm{F_{1}}$ &\textbf{Acc}	&\textbf{Prec}&\textbf{Rec}&  $\bm{F_{1}}$ \\
		\hline
		\multirow{5}{*}{CART} 
		& Textual&0.673&0.672&0.771&0.718&0.699&\textbf{0.647}&0.652&0.65\\
		& Image&0.632&0.701&0.639&0.668&0.647&0.595&0.533&0.563\\
		&MONITOR&\textbf{0.746}&\textbf{0.715}&\textbf{0.897}&\textbf{0.796}&\textbf{0.704}&0.623&\textbf{0.716}&\textbf{0.667}\\ \cline{2-10} 
		
		&Castillo&0.643&0.711&0.648&0.678&0.683&0.674&0.491&0.569\\	 
		&Wu&0.65&0.709&0.715&0.711&0.694&0.663&0.593&0.627\\
		
		\hline
		\multirow{5}{*}{KNN}
		& Textual&0.707&0.704&0.777&0.739&0.698&0.67&0.599&0.633\\
		& Image&0.608&0.607&0.734&0.665&0.647&0.595&0.533&0.563\\
		&MONITOR&\textbf{0.791}&\textbf{0.792}&\textbf{0.843}&\textbf{0.817}&\textbf{0.758}&\textbf{0.734}&\textbf{0.746}&\textbf{0.740}\\ \cline{2-10} 
		& Castillo&0.652&0.698&0.665&0.681&0.681&0.651&0.566&0.606\\	 
		&Wu&0.668&0.71&0.678&0.693&0.694&0.663&0.593&0.627\\
		\hline
		\multirow{5}{*}{SVM}
		& Textual&0.74&0.729&0.834&0.779&0.658&0.657&0.44&0.528\\
		& Image&0.693&0.69&0.775&0.73&0.595&0.618&0.125&0.208\\
		&MONITOR&\textbf{0.794}&\textbf{0.767}&\textbf{0.881}&\textbf{0.82}&\textbf{0.771}&\textbf{0.743}&\textbf{0.742}&\textbf{0.743}\\ \cline{2-10}
		& Castillo &0.702&0.761&0.716&0.737&0.629&0.687&0.259&0.377\\	 
		&Wu&0.725&0.763&0.73&0.746&0.642&0.625&0.394&0.484\\
		\hline
		\multirow{5}{*}{RF}
		& Textual&0.747&0.717&0.879&0.789&0.778&0.726&0.768&0.747\\
		& Image&0.652&0.646&0.771&0.703&0.652&0.646&0.771&0.703\\
		&MONITOR&\textbf{0.962}&\textbf{0.965}&\textbf{0.966}&\textbf{0.965}&\textbf{0.889}&\textbf{0.914}&\textbf{0.864}&\textbf{0.889}\\ \cline{2-10} 
		& Castillo&0.702&0.727&0.723&0.725&0.714&0.669&0.67&0.67\\	 
		&Wu&0.728&0.752&0.748&0.75&0.736&0.699&0.682&0.691\\
		\hline
	\end{tabularx}
\end{table}

\subsection{Classification Results}
From the classification results recorded in Table~\ref{tab8}, we can make the following observations.
\subsubsection{Performance Comparison} With MONITOR, using both image and textual feature allows all classification algorithms to achieve better performance than baselines. Among the four classification models, RF generates the best accuracy: 96.2\% on MediaEval and 88.9\% on FakeNewsNet, performing 26\% and 18\% better than Castillo and 24\% and 15\%  than Wu,  
still on MediaEval and FakeNewsNet, respectively. 

Compared to the 15 ``best'' textual feature set, RF improves the accuracy by more than 22\% and 10\% with image features only. Similarly, the other three algorithms achieve 
an accuracy gain between 5\% and 9\% on MediaEval and between 5\% and 6\% on FakeNewsNet.
Eventually, all classification algorithms generate a lower accuracy when using image features only.

While image features play a crucial role in rumor verification, we must not ignore the effectiveness of textual features. The role of image and textual features is complementary. When the two sets of features are combined, performance is significantly boosted.   

\subsubsection{Illustration by Example} To more clearly show the complementarity between text and images, we compare the results achieved with MONITOR and single modality approaches (text only or image only). Fake rumor messages from Figure~\ref{Fig1} (Section~\ref{sec:introduction}) are correctly detected as false by MONITOR, while using either only textual or only image modalities yields a true result.

In the tweet from Figure~\ref{a}, the text content solely describes the attached image without giving any signs about the veracity of the tweet. This is why the textual modality identifies this tweet as real. It is the attached image that looks quite suspicious. 
By combining textual and image contents, MONITOR can identify the veracity of the tweet with a high score, exploiting some clues from the image to get the right classification.

The tweet from Figure~\ref{b} is an example of rumor correctly classified by MONITOR, but incorrectly classified when only using the visual modality. The image seems normal and its complex semantics are very difficult to capture by the image modality. However, the words with strong emotions in the text indicate that it might be a suspicious message. By combining the textual and image modalities, MONITOR can classify the tweet with a high confidence score.

\subsection{Feature Analysis}
\begin{figure}[htp]
	\centering
	\subfloat[MediaEval\label{}]{%
		\includegraphics[width=0.28\textwidth]{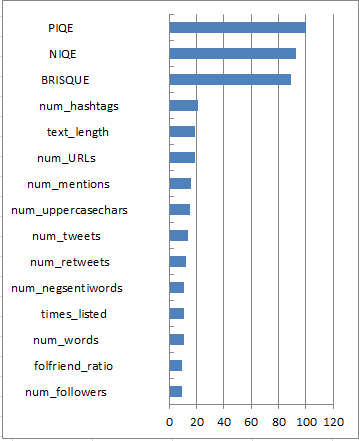}%
	}\hfil
	\subfloat[FakeNewsNet\label{}]{%
		\includegraphics[width=0.25\textwidth]{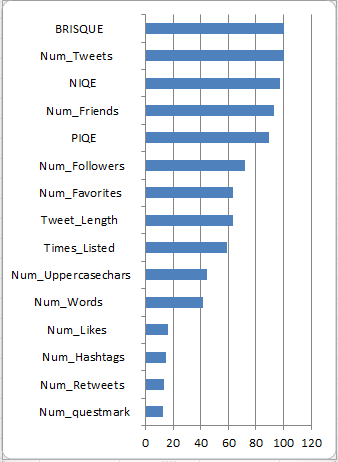}%
	}
	\caption{Random Forest feature importance}
	\label{Fig5}
	
\end{figure}

The advantage of our approach is that we can achieve some elements of interpretability. To this aim, we conduct an analysis to illustrate the importance of each feature set. We depict the first most 15 important features achieved by RF in 
Figure~\ref{Fig5}, which shows that, for both datasets, visual characteristics are in the top-five features. The remaining features are a mix of text content and social context features. These results validate the effectiveness of the IQA image features, as well as the the importance of fusing several modalities in the process of rumor verification.

\begin{figure}[htp]
	\centering
	\subfloat[MediaEval\label{}]{%
		\includegraphics[width=0.50\textwidth]{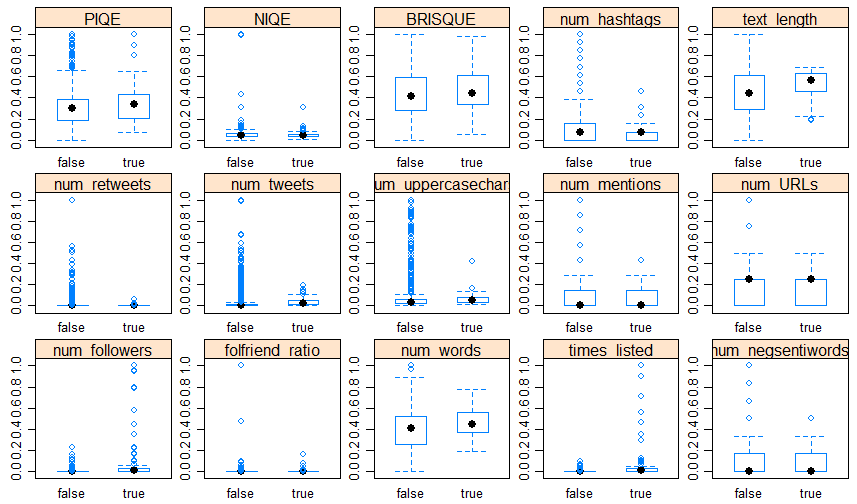}%
	}\quad
	\subfloat[FakeNewsNet\label{}]{%
		\includegraphics[width=0.46\textwidth]{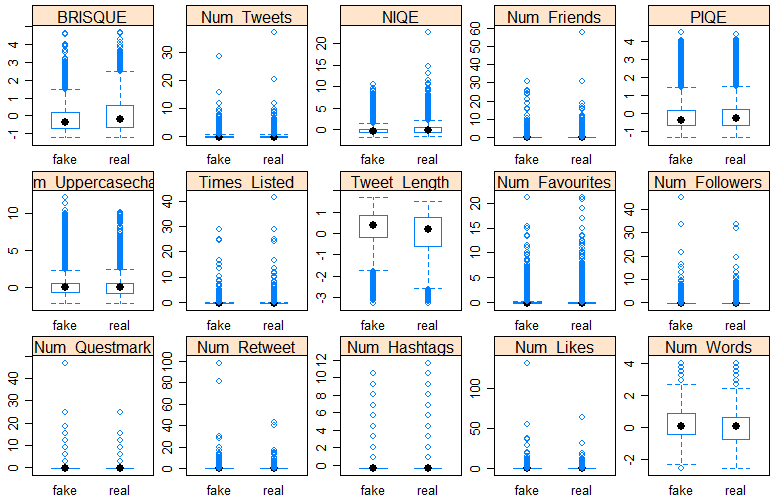}%
	}
	\caption{Distribution of true and false classes for top-15 important features}
	\label{Fig6}
	
\end{figure}

Eventually, to illustrate the discriminating capacity of these features, we deploy box plots for each of the 15 top variables on both datasets. Figure~\ref{Fig6} shows that several features exhibit a significant difference between fake and real classes, which explains our good results.

\subsection{Early and Late Fusion}

In our previous experiments, we fuse visual and textual modalities into a single multimodal vector before the learning and classification steps, in the so-called \textit{early fusion} manner. Another way to merge features is \textit{late fusion}.

This class of fusion scheme works at the decision level, by combining the prediction scores available for each modality. Late fusion starts with the extraction of unimodal features. In contrast to early fusion, where features are combined into a multimodal representation, late fusion approaches learn directly from unimodal features. The predicted probability scores are combined afterwards to yield a final detection score. Several methods help combine scores, such as averaging, voting or using another machine learning method to learn how to best combine predictions. 

\begin{figure}[htp]
	\centering
	\includegraphics[width=0.9\textwidth]{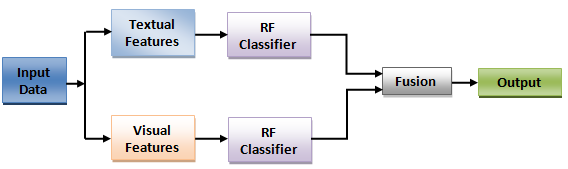}
	\caption{Late fusion scheme} \label{Fig7}
\end{figure}

To apply late fusion, we train two Random Forest (RF) classifiers by learning separately the visual and textual features (Figure~\ref{Fig7}). 

To obtain the final classification results, the predicted probabilities of the both classifiers are combined with (1) equal weights, by assuming that the two models are equally skillful and make the same proportional contribution to the final prediction; and (2)  averaging the (optimized) weights by feeding the classifiers' output to a logistic regression model.

\begin{figure}[htp]
	\centering
	\subfloat[MediaEval\label{}]{%
		\includegraphics[width=0.45\textwidth]{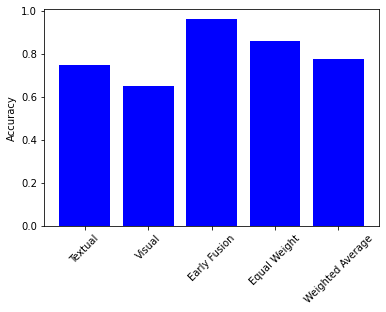}%
	}\vspace{0.5cm}
	\subfloat[FakeNewsNet\label{}]{%
		\includegraphics[width=0.45\textwidth]{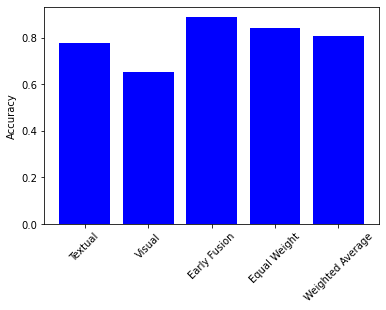}%
	}
	\caption{Performance of early and late fusion}
	\label{Fig8}
\end{figure}

Figure \ref{Fig8} shows that, for both datasets, the early fusion method and the  two late fusion strategies, i.e., equal weight and optimized weight, boost the  prediction with different rates using separately two sets of features. Early fusion has the highest performance score, while for both late fusion techniques, equal weight is slightly more efficient than optimized weight. 

Late fusion's performance is lower than that of early fusion because, when we train two models separately on visual and textual features, some dependencies between features are lost. Practically, there are some correlations between features, e.g., between BRISQUE and Num\_Mention or between PIQE and Text\_Length. The potential loss of correlation in the mixed feature space 
is a drawback of late fusion. Another disadvantage of late fusion is its cost in terms of learning effort, as every modality requires a separate supervised learning stage. Moreover, the combined representation requires an additional learning stage.

\section{Ensemble Learning Performance}
\label{sec:ensemble}

Applied machine learning often involves fitting and evaluating models on a dataset. Given that we cannot know what model will perform best on the dataset beforehand, this may involve a lot of trial and error until we find a model that performs good enough. This is akin to making a decision using the single expert we can find. A complementary approach is to prepare multiple, different models, and then combine their predictions using an ensemble machine learning model.

\begin{figure}[htp]
	\centering
	\includegraphics[width=0.7\textwidth]{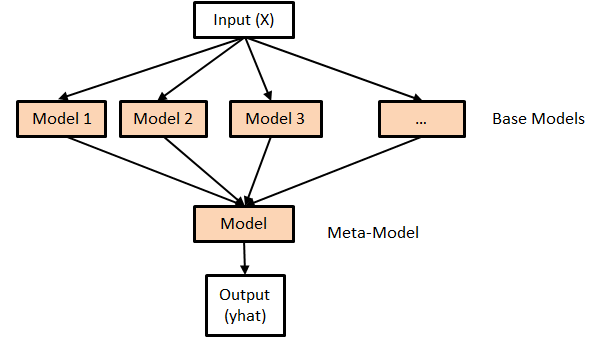}
	\caption{Stacking ensemble} \label{Fig9}
\end{figure}

Because ensemble learning strategies such as bagging and boosting typically involve a single machine learning algorithm (generally a decision tree), we use instead the stacking strategy (also called metalearning) that seeks for a diverse group of members by varying model types. Figure \ref{Fig9} summarizes the key elements of a stacking ensemble:
\begin{itemize}
	\item an unchanged training dataset;
	\item various machine learning algorithms (base models) for each ensemble member; 
	\item a machine learning model (metamodel) to learn how to best combine predictions.
\end{itemize}

To measure the performance of ensemble learning models for rumor detection, we develop five metamodels as variants of the stacking strategy.

\subsection{Metamodels}

\subsubsection{Voting Ensemble} We construct two voting models. The first one is a soft voting model called MONITOR$_{sv}$ that sums  the predictions made by the classification models listed in Table \ref{tab8} and predicts the class label with the largest sum probability. 
The second model is a weighted average voting model called MONITOR$_{wav}$ where model votes are proportional to model performance. The performance of each ensemble model on the training dataset will be used as the relative weighting of the model when making predictions. Performance is calculated using classification accuracy as a ratio of correct predictions ranging between 0 and 1, with larger values meaning a better model and, in turn, more contribution to the prediction.

\subsubsection{Canonical Stacking Ensemble} 

Following \cite{wolpert1992stacked}'s canonical stacking strategy (Figure \ref{Fig8}), we construct a model called MONITOR$_{st}$. Concretely, we use three repeats of a stratified 10-fold cross-validation on the four classification models to prepare the training dataset (predictions) with the logistic regression metamodel. Furthermore, we train the metamodel on the prepared dataset as well as the original training dataset using a 5-fold cross-validation. This aims to provide an additional context to the metamodel to better combine predictions. 

\subsubsection{Blending ensemble} 
Blending was the term commonly used for stacking ensembles during the Netflix prize in 2009. The prize involved teams seeking movie recommendations that performed
better than the native Netflix algorithm. A one million US dollar prize was awarded to the team achieving a 10\% performance improvement.

In this stacking-type ensemble, base models are fit on the training dataset and the metamodel is trained on predictions made by each base model on the validation dataset. At the time we are writing this paper, Scikit-learn does not  support blending. Thus, we implement a blending model called MONITOR$_{bld}$ using scikit-learn models. 

\begin{figure}[htp]
	\centering
	\includegraphics[width=0.7\textwidth]{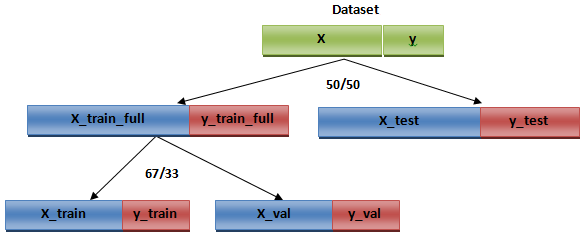}
	\caption{Dataset splitting} \label{Fig10}
\end{figure}

To implement our model, we need to split the dataset, first into training and test sets. Then, the training set is split again into two subsets used to train base models and the metamodel, respectively. We use a 50/50 split on the training and test sets and  a 67/33 split on the train and validation sets (Figure~\ref{Fig10}). Furthermore, we choose logistic regression as a metamodel (the blender), for the same reasons we mentioned about canonical stacking. We summarise the key implementation steps of our model in Algorithm \ref{alg:PL}.

\begin{algorithm}[htp]

	\caption{Blending Ensemble}\label{alg:PL}
	\begin{algorithmic}[1]
	    \Require $Dataset(\Var{X,y})$ \Comment{input variables and output label}
	    \State \Var{meta_x, meta_y}  $\leftarrow$ \Var{empty list}
		\State{Split \textit{Dataset} into \Var{X_train, y_train}, \Var{X_val, y_val} and \Var{X_test, y_test}} \Comment{train, validation and test sets}
		\State{Create base models}
		\LineComment{Fit the blending ensemble}
		\ForAll{\text{base-model}}
		    \State Fit base-model on training set (\Var{X_train, y_train})
		    \State Predict with base-model on \Var{X_val}
		    \State Store predictions in \Var{meta_x} 
	    \EndFor
	    \State Convert \Var{meta_x} to 2D array \Comment{as input for blending model} 
	    \State Define blending model
		\State Fit blending model on predictions from base models (\Var{meta_x,y_val})
		\LineComment{Make prediction with blending ensemble}
		\ForAll{\text{base-model}}
		    \State Predict with base-model on \Var{X_test}
		    \State Store predictions in \Var{meta_y} 
	    \EndFor
	    \State Convert \Var{meta_y} to 2D array \Comment{as input for blending model}
		\State Predict with blending model on \Var{meta_y} 
		\State Evaluate blending model on \Var{y_test}  
	\end{algorithmic}
\end{algorithm}

\subsubsection{Super Learner Ensemble} 
\begin{figure}[htp]
	\centering
	\includegraphics[width=0.9\textwidth]{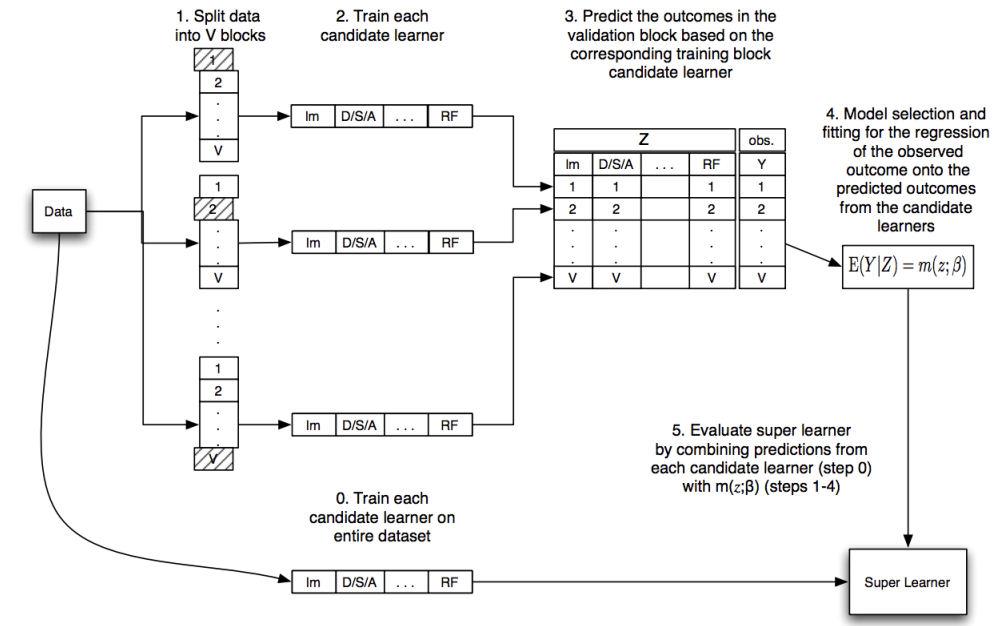}
	\caption{Super learner ensemble data flow \citep{van2007super}} \label{Fig11}
\end{figure}

A super learner ensemble \citep{van2007super} is a specific stacking configuration where all base models use the same $k$-fold splits of data, and a metamodel is fit on the out-of-fold predictions from each model. We summarize this procedure in Algorithm \ref{alg:SL}. Moreover, Figure \ref{Fig11}, which is reproduced from the original paper by \citep{van2007super}, depicts its data flow. We use the MLENS Python library~\citep{flennerhag:2017mlens} to implement the super learner model called MONITOR$_{sl}$, where we split the training data into $k = 10$ folds. The number of base models is set to $m = 4$(i.e. KNN, CART, SVM and RF).

\begin{algorithm}[htp]
	\caption{Super learner ensemble}\label{alg:SL}
	\begin{algorithmic}[1]
		\State{Select a $k$-fold split of the training dataset}
		\State{Select $m$ base-models or model configurations} 
		\ForAll{\text{base-model}}
		    \State Evaluate using $k$-fold cross-validation
		    \State Store all out-of-fold predictions 
		    \State Fit the model on the full training dataset and store
		\EndFor
		\State Fit a metamodel on the out-of-fold predictions
		\State Evaluate the model on a holdout dataset or use model to make predictions
	\end{algorithmic}
\end{algorithm}

Table \ref{tab9} summarizes the results achieved by the best individual machine learning model (RF) and the five stacking algorithms.

\begin{table}[h]
	\caption{Performance of MONITOR and stacking ensemble models}\label{tab9}
	\centering
	\begin{tabularx}{\textwidth}{bssssssss}
		\hline
		&\multicolumn{4}{l}{\textbf{MediaEval}}&\multicolumn{4}{l}{\textbf{FakeNewsNet}}\\
		\cline{2-3} \cline{6-7}
		\textbf{Model} & & & & & & & & \\
		&\textbf{Acc} & \textbf{Prec}&\textbf{Rec}&  $\bm{F_{1}}$ &\textbf{Acc}&\textbf{Prec}&\textbf{Rec}&  $\bm{F_{1}}$ \\
		\hline
		MONITOR&0.962&0.965&0.966&0.965&0.889&0.914&0.864&0.889\\
		\hline
		MONITOR$_{sv}$ &0.966&0.955&0.976&0.965&0.897&0.911&0.873&0.892\\
		\hline
		MONITOR$_{wav}$&0.968&0.968&0.970&0.969&0.906&0.90&0.927&0.914\\
		\hline
		\textbf{MONITOR}$\bm{_{st}}$&\textbf{0.984}&\textbf{0.979}&\textbf{0.989}&\textbf{0.984}&\textbf{0.936}&\textbf{0.929}&\textbf{0.952}&\textbf{0.941}\\
		\hline
		MONITOR$_{bld}$&0.973&0.975&0.971&0.973&0.915&0.909&0.932&0.921\\
		\hline
		MONITOR$_{sl}$&0.970&0.980&0.959&0.969&0.921&0.915&0.937&0.926\\
		\hline
	\end{tabularx}
\end{table}

\subsection{Result Analysis}
Our comparative analysis of experimental results shows that all metalearning models are more efficient than the best individual machine learning model (RF), because by combining multiple models, the errors from a single base-model are likely  compensated by the other models. As a result, the overall prediction performance of the ensemble is better than that of any single base-model.  

\begin{figure}[htp]
	\centering
	\subfloat[Accuracy\label{}]{%
		\includegraphics[width=0.5\textwidth]{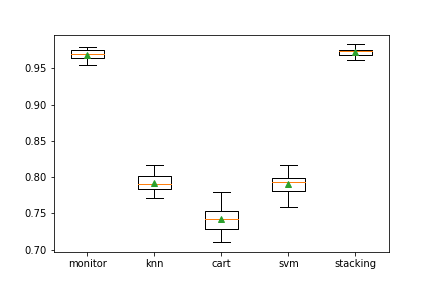}%
	}\vspace{0.5cm}
	\subfloat[ROC\label{}]{%
		\includegraphics[width=0.45\textwidth]{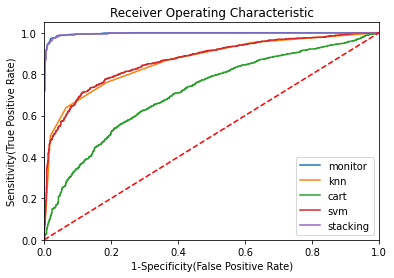}%
	}
	\caption{Stacking ensemble model vs. standalone models on MediaEval}
	\label{Fig12}
\end{figure}

\begin{figure}[htp]
	\centering
	\subfloat[Accuracy\label{}]{%
		\includegraphics[width=0.5\textwidth]{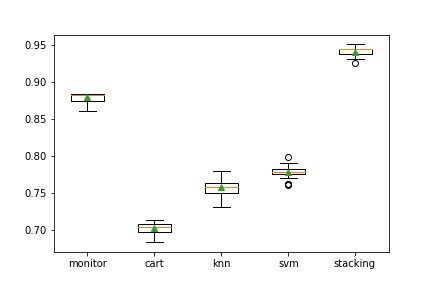}%
	}\vspace{0.5cm}
	\subfloat[ROC\label{}]{%
		\includegraphics[width=0.45\textwidth]{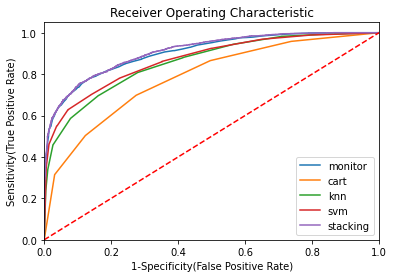}%
	}
	\caption{Stacking ensemble model vs. standalone models on  FakeNewsNet}
	\label{Fig13}
\end{figure}

Moreover, for both datasets, the canonical stacking algorithm outperforms all models with 98.4\% and 93.6\% of accuracy on MediaEval and FakeNewsNet dataset, respectively. The stacking model indeed takes advantages from the diversity of  predictions made by contributing models. That is, all algorithms are skillful on the classification problem, but in different ways. Figures~\ref{Fig12} and~\ref{Fig13} depicts the accuracy score box plot and the Receiver Operating Curve (ROC) for the canonical stacking ensemble model compared to the standalone machine learning algorithms (MONITOR-RF, CART, KNN and SVM) on MediaEval and FakeNewsNet, respectively.

Among the five ensemble models, the soft voting algorithm achieves the worst results, because 
it treats all models the same, i.e., all models contribute equally to the prediction. 
Although the canonical stacking algorithm performs the best, the blending and super learner algorithms achieve scores that are very close to those of stacking and therefore turn to be useful too for rumor classification.

\section{Conclusion and Perspectives}
\label{sec:conclusion}
To assess the veracity of messages posted on social networks, most of the existing
techniques ignore visual contents and use traditional machine learning models for classification, although ensemble approaches are considered the state-of-the-art solutions for many machine learning challenges. Thence, in this paper, to improve the performance of message verification, we propose a multimodal fusion framework called MONITOR that uses features extracted from the textual content of messages, the social context and image features that have not been considered until now. We compare the performance of MONITOR with five metalearning ensemble models by combining four base-predictors (KNN, CART, SVM and RF).
Extensive experiments conducted on the MediaEval benchmark and the FakeNewsNet dataset show that: 
\begin{itemize}
    \item the image features that we introduce play a key role in message veracity assessment;
    \item no single homogeneous feature set can generate the best results alone;
    \item all ensemble algorithms outperform the best single base-model (RF), and canonical stacking achieves the best performance on both datasets.
\end{itemize}

Our future research includes two directions. In the short term, we plan to experiment with other, larger datasets
and vary the type, combination and number of base
models in the ensemble. Second, we plan to compare MONITOR's performance with a deep learning-based approach for rumor classification, deepMONITOR \citep{azri2021calling}, with the aim of studying the tradeoff between classification accuracy, computing complexity and explainability.

\section{Declarations}
\paragraph{Conflict of Interest} The authors have no conficting interests to declare that are relevant to the content of this article.

\bibliography{sn-bibliography}


\section*{Authors' Biographies}

\textbf{Abderrazek Azri} received the Master’s degree in information systems security from the University of Lyon~2, Lyon, France, in 2014, he received the Ph.D. degree from the University of Lyon~2, Lyon, France, in july 2022. His research interests include social media analysis and data mining.
\vspace{0.4cm}

\noindent
\textbf{Cécile Favre} is a lecturer in computer science at the University of Lyon 2 and a member of the ERIC laboratory. She is also an associate researcher at the Centre Max Weber. His work focuses on the one hand on computer science research (decision-making computing, social network analysis with in particular work in the field of bibliometrics). On the other hand, she is developing interdisciplinary work around gender issues and IT. She is also involved in various responsibilities in the Mention of Master in Gender Studies in Lyon.
\vspace{0.4cm}

\noindent
\textbf{Nouria Harbi} is a member of research staff ERIC Laboratory, University Lyon~2, France. She received her master degree in computer science and Ph.D. from INSA Lyon, France. She then joined the laboratory ISEOR, where she worked on information systems. She is currently working on the security of decisional information system and modelling data warehouse.
\vspace{0.4cm}

\noindent
\textbf{Jérôme Darmont} is full professor of computer science at the University of Lyon, France. He received his Ph.D. in 1999 from the University of Clermont-Ferrand II, France, and then joined the University Lyon~2 as an associate professor. He became full professor in 2008 and has been director of the ERIC research center from 2012 to 2021. In 2017, he was made Honoris Causa professor at Simon Kuznets Kharkiv National University of Economics, Ukraine. He is currently adjunct director of the Institute of Communication at University Lyon~2. His research interests mainly relate to data management performance (performance optimization, auto-administration and benchmarking of databases, data warehouses, data lakes, data meshes...) and cloud business intelligence (data security, query performance and cost, personal BI, big data analytics, textual document analysis...). 
\vspace{0.4cm}

\noindent
\textbf{Camille Nôus} came into existence on 20 March 2020, to represent the contribution of the academic community to research in France, in the form of a collective and gender-neutral signature. This signature, devised as a scientific consortium, calls for an open and collaborative approach to the creation and diffusion of knowledge, under the aegis of the academic community, and is intended to be a mark of integrity.
\end{document}